\documentclass{article}

\usepackage[preprint, nonatbib]{neurips_2020}

\usepackage[utf8]{inputenc} % allow utf-8 input
\usepackage[T1]{fontenc}    % use 8-bit T1 fonts
\usepackage{hyperref}       % hyperlinks
\usepackage{url}            % simple URL typesetting
\usepackage{booktabs}       % professional-quality tables
\usepackage{amsfonts}       % blackboard math symbols
\usepackage{nicefrac}       % compact symbols for 1/2, etc.
\usepackage{microtype}      % microtypography
\usepackage{amsmath}
\usepackage{cmap}
\usepackage{graphicx}
\usepackage{amsmath}
\usepackage[numbers]{natbib}
\title{Tensorized Transformer for Dynamical Systems Modeling}

\author{%
  Anna Shalova \\%\thanks{footnote} \\
  Skolkovo Institute of Science and Technology \\
  Moscow, Russia \\
  \texttt{anna.shalova@skoltech.ru} \\
 \And
  Ivan Oseledets \\
  Skolkovo Institute of Science and Technology \\
  Moscow, Russia \\
  \texttt{i.oseledets@skoltech.ru} \\
}

\begin{document}

\maketitle

\begin{abstract}
The identification of nonlinear dynamics from observations is essential for the alignment of the theoretical ideas and experimental data. The last, in turn, is often corrupted by the side effects and noise of different natures, so probabilistic approaches could give a more general picture of the process. At the same time, high-dimensional probabilities modeling is a challenging and data-intensive task. In this paper, we establish a parallel between the dynamical systems modeling and language modeling tasks. We propose a transformer-based model that incorporates geometrical properties of the data and provide an iterative training algorithm allowing the fine-grid approximation of the conditional probabilities of high-dimensional dynamical systems.
\end{abstract}

\section{Introduction}

Dynamics modeling of complex systems is one of the key problems in different areas of science. The underlying dependencies rising from the real-life systems are often complex or even unpredictable. Numerical modeling of their solutions is a challenging task of computational science. The main factor complicating the dynamical systems modeling is the sensitivity to initial conditions. If the solutions of the system with arbitrary close initial conditions exponentially diverge, then even small error in predictions leads to complete divergence of the approximation model.

In this paper, we are focusing on the problem of the dynamical systems modeling from observations. We are given a set of trajectories of the dynamical system:
\begin{equation}\label{dynGPT-2:sys}
  \frac{dx}{dt} = f(x), \quad x(0)=x_0, \quad x \in \mathbb{R}^d,
\end{equation}
and our goal is to train a model reproducing the underlying dynamics. We focus on the specific class of dynamical systems, deterministic chaos, combing sensitivity to initial conditions with the existence of a bounded invariant set. Even though the solutions of such a system are bounded, they exponentially diverge from each other:
\begin{equation}
    \|x_1(t) - x_2(t)\| \sim e^{\lambda t}\|x_1(0) - x_2(0)\|.
\end{equation}
This complex structure of an attractor is caused by the properties of the underlying equations. Therefore, if we directly approximate function $f(x)$, even a small deviation from the functional dependence results in the trajectories leaving the invariant set. As a result, a straightforward approximation of the function $f(x)$ by a neural network does not allow preserving the properties of the system. 

To overcome this issue, we introduce a scalable probabilistic approach to the dynamical systems modeling based on the transformer architecture. The method directly exploits the geometry of the phase space, which allows modeling of the high-dimensional systems. The model has $O(Nd)$ parameters, where $N$ is the size of the grid and $d$ - dimensionality of the dynamical system. %We also propose an iterative training algorithm allowing the fine-grid approximation of the high-dimensional systems. 

Main contributions of this paper are:
\begin{enumerate}
    \item We propose a geometry-aware transformer-decoder model for the dynamical systems modeling that incorporates the factorization of both embedding and classification layers.
    \item We introduce an iterative training approach that significantly speeds up the convergence of the model. 
    \item We apply the model to the simulated data and demonstrate that it outperforms baselines on the Rossler system and shows competitive performance on the Lorenz-96 systems. 
\end{enumerate}

\section{Proposed Method}
%We consider the standard setup of the dynamics identification from the experimental data. We are given a set of the solutions of a system \ref{dynGPT-2:sys} $x^{(i)}(t_k), i = 1, \ldots, M, k = 1, \ldots, T$, and our goal is to get a model that is able to reproduce the underlying dynamics. In case of the continuous time system \ref{dynGPT-2:sys}, the trajectories are firstly discretized in time, such that $x(t_k) = x(k\tau)$, where $\tau$ is the constant time step. Trajectories of this form can represent both simulated and experimental data. Then, given a beginning of some trajectory $x_1, \ldots, x_K$ the model is expected to predict the next state $x_{K+1} = x(t_{K+1})$.
\begin{figure}
    \centering
    \includegraphics[width=0.8\textwidth]{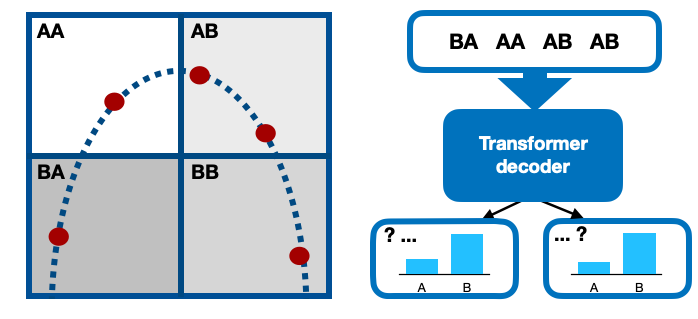}
    \caption{An example of the proposed method for the 2-dimensional trajectory. The initial states are discretized along each dimension; then transformer decoder is used to predict conditional probabilities along each dimension.}
\end{figure}

For chaotic systems, the accuracy of the long-term predictions is a questionable functional, since even for the original system, small perturbation leads to large errors in the trajectories. Instead, we can approximate the stable quantities of dynamical systems, such as the invariant measure. In order to do it, instead of dealing with continuous space-time, we discretize them both and convert the dynamical system modeling into discrete sequence modeling. 

The goal is now to predict the conditional probability distribution.
The main challenge is that the phase-space is high-dimensional, and discretization with a uniform grid requires $O(N^d)$ "words", making this approach impractical for $d > 3$ due to the curse of dimensionality. However, such discretization is perfect for the use of \emph{tensor factorization methods}. Standard matrix of embeddings for the given vocabulary would have the shape $N^d \times R$, where $R$ is the size of representations. To avoid the exponential growth of the size of this matrix with respect to $d$, we introduce the \emph{tensor coding layer}, which factorizes these embeddings into  \emph{geometry-aware} components. 
Similarly, we handle the classification layer, reducing a single classification problem to the separate prediction of the indexes along each of the $d$ axis. To get a trajectory in a continuous space we map predicted indexes back to the centers of the corresponding discretization segments.

\begin{figure}
    \centering
    \includegraphics[width=\textwidth]{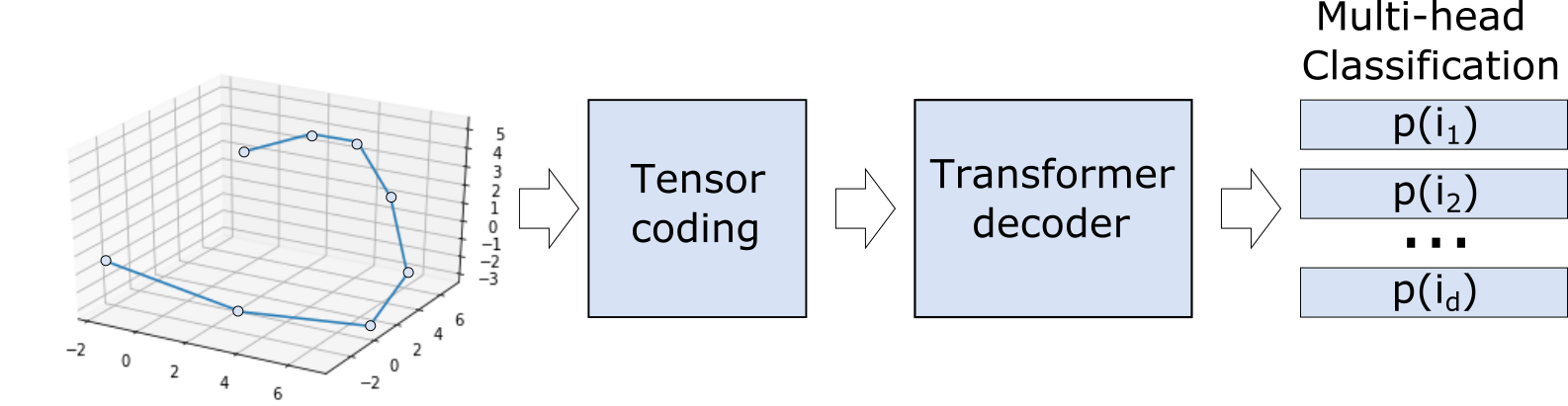}
    \caption{General scheme of the model: firstly, the states are mapped to the discrete space, where the representations are calculated with tensor coding layer; then the dynamics of these representations is modeled with a transformer decoder; and finally, the factorized probabilities of the next state are predicted with a multi-head classification layer.}
    \label{fig:scheme}
\end{figure}
\begin{figure}
    \centering
    \includegraphics[width=\textwidth]{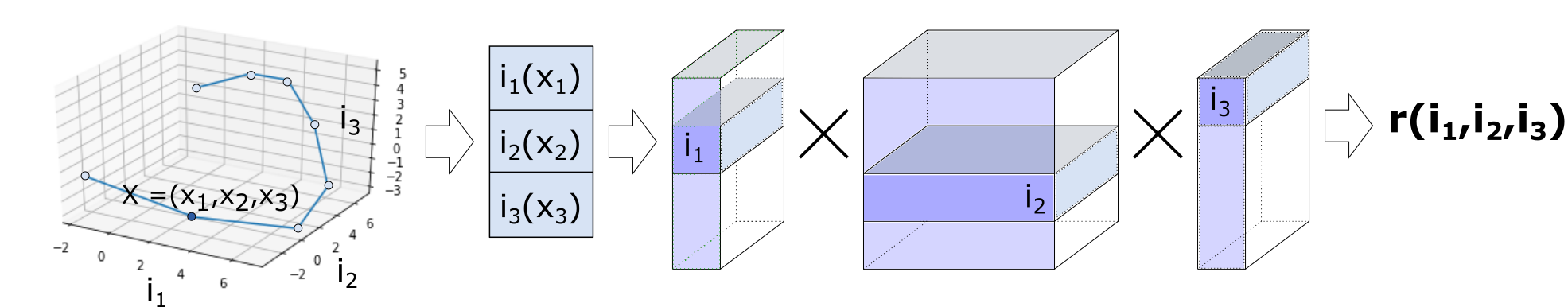}
    \caption{An example of the tensor coding layer: continuous point $X = (x_1, x_2, x_3)$ is discretized along each dimension ($i_1, i_2, i_3$), then the representations $r(i_1, i_2, i_3)$ are calculated using the corresponding components of the TT-embeddings.}
    \label{fig:coding}
\end{figure}
%However, our task is to predict the next symbol, which is also a high-dimensional, so we propose factorizing conditional probabilities in order to reduce size of the classification layer. 

\subsection{Factorized probabilistic model}
We are given a set of the solutions of a system \ref{dynGPT-2:sys}: $x^{(i)}(t_k), i = 1, \ldots, M, k = 1, \ldots, T$, such that $x(t_k) = x(k\tau)$, where $\tau$ is the constant time step. In real life, trajectories of the dynamical system are measured with a limited accuracy. This uncertainty can be viewed as an additive noise. So instead of the maximum-likelihood estimation of the next state:
\begin{equation}
    x_{K+1} = F(x_1, \ldots, x_K),
\end{equation}
we propose to model the conditional distribution:
\begin{equation}
   p(x_{K+1}\vert x_1, \ldots, x_K).
\end{equation}
For the discrete $x_k$, this task is known as \emph{language modeling}. 
Similar to the dynamical systems, for a long time, language modeling tasks were being solved by recurrent neural networks \cite{kalchbrenner2013recurrent, sutskever2014sequence, cho2014properties, bahdanau2014neural}, until the transformer model has revolutionized natural language processing \cite{vaswani2017transformer}. Self-attentive models remain the state-of-art architectures in most of the NLP tasks \cite{vaswani2017transformer, devlin2018bert, radford2019gpt, brown2020language} and lately are also adopted to the tasks from other areas of science \cite{sun2019contrastive, carrasquilla2019quantum, zhang2019self}. 

In this paper, we build upon methodology proposed in \cite{shalova2020deep}: we reduce the task of dynamical systems modeling to the language modeling using an equidistant grid discretization of the trajectories and apply transformer decoder to model dynamics in the discrete space. Each state of $d$-dimensional system $x$ is mapped to the multi-index representation $\{i_1, \ldots i_d\}, \ i_k \in \{1\ldots N \}$, where $N$ is the grid size and each index indicates the position along one of the dimensions. High-dimensional probabilities modeling is a very challenging task even in the discrete space case, so instead of the direct modeling of the $d$-dimensional $p(x_{K+1}), \ x_{K+1} \in \mathbf{R}^d$, we propose to approximate it with a factorization into the one-dimensional distributions corresponding to the different dimensions:

\begin{equation}
\begin{split}
    p(x_{K+1}) = p(x_{K+1}^1)\times\ldots \times p(x_{K+1}^d), 
    \\
    p(x_{K+1}^i) = p(x_{K+1}^i|x_1, \ldots, x_K).
\end{split}
\end{equation}
Such factorization makes easy both the likelihood computation and the sampling from the predicted distribution.

 In other words, the indices $\{i_1, \ldots i_d\}$ are modeled by \emph{separate classification heads}, that are learned to maximize the KL-divergence between empirical conditional distributions of the training set and predicted factorized distribution (Figure \ref{fig:scheme}). Each of $d$ heads has $O(N)$ parameters, which gives an $O(dN)$ parameters for the classification layer size. 

\subsection{Tensor coding}
At the same time, each component of the distribution is conditioned by the whole information about previous states. It requires training of the unique representation for each of the $N^d$ unique states. Similar to the factorization of the probabilities, we propose the usage of the factorized embedding layer that relies on the known relation between the states. As such factorization, we use a tensor coding layer, a geometry-aware analog of the tensor-train embeddings \cite{khrulkov2019ttembeddings}. With a fixed representation size, it gives a $O(\log N^d)= O(d)$ estimation of the number of the parameters. In \cite{khrulkov2019ttembeddings}, it was proposed to train the embeddings matrix in the format of a tensor-train matrix. Instead of matrix $M: V \times D$, where $V$ is the size of vocabulary and $D$ is a representation size, there is a $k$-dimensional tensor $A: v_1 \times v_2 ... \times d_1 \times d_2 ... $, such that $\prod_j v_j = V, \ \prod_i d_i = D$. According to \cite{oseledets2011tensor}, tensor $A$ can be approximated by the tensor-train decomposition:
\begin{equation}
    \label{eq:tt}
    A(i_1, i_2, ... i_k) = \sum_{\alpha_0, \alpha_1, ...\alpha_d}G_1(\alpha_0, i_1, \alpha_1) G_2(\alpha_1, i_2, \alpha_3)...G_d(\alpha_{d-1}, i_d, \alpha_d),
\end{equation}
where $\alpha_i$ are the ranks of the decomposition. In case of the embeddings layer in deep neural networks, during training, instead of the approximation of the full-rank matrix $M$, we optimize parameters of the decomposition $G_i$. In our case, we propose using a single component of the TT-decomposition for every dimension. That additionally allows an intuitive restoration of the chosen embeddings without computation of the whole matrix.

\section{Iterative training}
\begin{figure}
    \centering
    \includegraphics[width=0.49\textwidth]{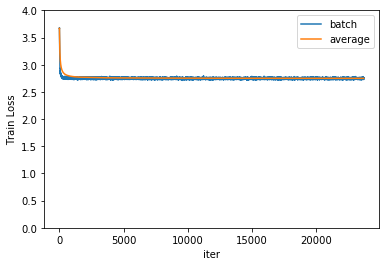}
    \includegraphics[width=0.49\textwidth]{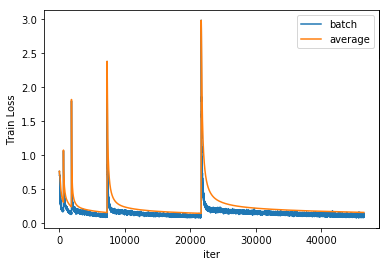}
    \caption{Training loss for the standard procedure (left) and with the iterative training algorithm (right). The peaks corresponds to the transition to the finer grids.}
    \label{fig:training}
\end{figure}

We found that direct training of the tensorized transformer-decoder model does not lead to good convergence (Figure \ref{fig:training}). However, uniform grid discretization introduces a natural hierarchy of models, and we can use the idea of a multigrid method from numerical analysis. Similar approaches have been successfully used in machine learning in progressive growing GANs, which were iteratively trained starting from a low resolution \cite{karras2017progressive, acharya2018towards}.

Each component of the embeddings factorization can be interpreted as an additional layer in the model. So, for high-dimensional systems, we have a deep architecture with a slow convergence. Instead of the direct optimization of the model with a desirable discretization, we propose training model iteratively such that, on each step, the discretization scale is decreased. We begin with training the transformer decoder to reproduce dynamics on the roughest possible discretization with $N=2$. Then, on each step, we use pre-trained embeddings corresponding to $N$ segments to initialize embeddings with discretization size $2N + 1$ and fine-tune both representations and dynamics model on the obtained finer grid. 

Each component of the TT-embeddings is a 3-dimensional tensor $G_k:\  R \times N \times R$, where $R$ is the rank of decomposition and $N$ is the discretization size. To initialize the $k$-th component of the finer-grid model $\tilde{G}_k: \ R \times 2N+1 \times R$ we use linear approximation of the $G_k$ tensor. The even segments are initialized directly with the components of $G_k$, and the odd components are the average between the adjacent ones:
\begin{equation}
\begin{split}
    \tilde{G}_k(:, 2i, :) = G_k(:, i, :), \ i=0, 1 ...N, 
    \\
    \tilde{G}_k(:, 2i+1, :) = \frac{G_k(:, i, :) + G_k(:, i+1, :)}{2}, \ i=0, 1 ...N-1.
\end{split}
\end{equation}
In section \ref{experiment:lorenz}, we demonstrate that linear approximation is appropriate as the learned representations are smooth along each direction.

On each fine-tuning iteration the model is trained to maximize the log-likelihood of the discrete trajectories. Easy to notice, that the factorization of the distribution leads to the factorization of the loss function:
\begin{equation}
\begin{split}
    \mathcal{L} = -\frac1{N}\sum_t\log p(x_t|x_1, \ldots x_{t - 1}) = -\frac1{N}\sum_t\sum_i \log p(x_t^i|x_1, \ldots, x_{t-1})  = \\
    = \sum_i \big(-\frac1{N}\sum_t \log p(x_t^i|x_1, \ldots, x_{t-1})\big),
\end{split}
\end{equation}
which corresponds to the standard multi-head classification problem. 

\section{Experiments}
\begin{figure}
    \centering
    \includegraphics[width = 0.49\textwidth]{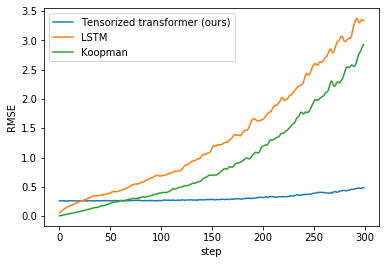}
    \includegraphics[width = 0.49\textwidth]{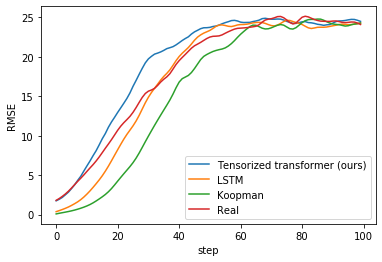}
    \caption{Average over 100 trajectories error of predictions from time for LSTM, Deep Koopman and Tensorized Transformer (ours) models for: (right) Rossler attractor and (left) Lorenz-96 system with $d = 16$. For Rossler we use grid with $N = 50$ segments, for Lorenz-96 - $N = 33$. For the Lorenz-96 system we also show the divergence of the real trajectories.}
    \label{fig:rmse}
\end{figure}
In all experiments we use Hugginface implementation\footnote{\url{https://github.com/huggingface/transformers}} of the transformer decoder in PyTorch. Our implementation of the TT-embeddings is based on the code\footnote{\url{https://github.com/KhrulkovV/tt-pytorch}} for \cite{khrulkov2019ttembeddings} except that on the forward pass all representations are reconstructed separately without restoration of the whole embeddings matrix. For both systems we use TT-embeddings with the rank $16$. Training data is simulated using an \emph{odeint} integrator from the SciPy library in Python.

In all cases, we use a zero-temperature generation, meaning that on each step model selects the mode of the predicted distribution. For the used dynamical systems, the learned distributions were shown to be close to the delta functions, which means that the characteristic size of the distribution was smaller than the size of discretization segments. In such a case, higher-temperature generation, or, in other words, random index selection with the scaled predicted probabilities, is noisy because of the specific distribution form.

As baselines we use LSTM \cite{hochreiter1997long} and Deep Koopman \cite{lusch2018deepkoopman} models. For both dynamical systems, we use a single-layer LSTM model with a hidden size $256$. In the case of the Deep Koompan model, both encoder and decoder consist of three linear layers followed by $tanh$ activation; we set the hidden size = $512$. 

\begin{figure}
    \centering
    \includegraphics[width=0.329\textwidth]{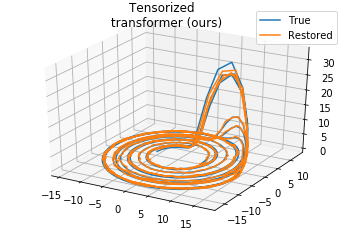}
    \includegraphics[width=0.329\textwidth]{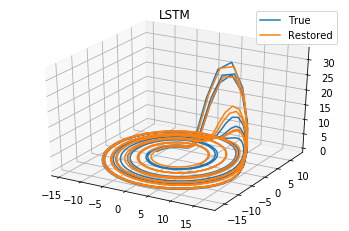}
    \includegraphics[width=0.329\textwidth]{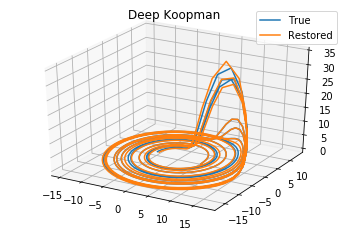}
    \caption{Trajectory of the Rossler attractor restored by LSTM, Deep Koopman and Tensorized Transformer model with $N = 50$ (ours). Both LSTM and Transformer are conditioned on 100 states of the real trajectory.}
    \label{fig:rossler}
\end{figure}
\begin{figure}
    \centering
    \includegraphics[width=0.9\textwidth]{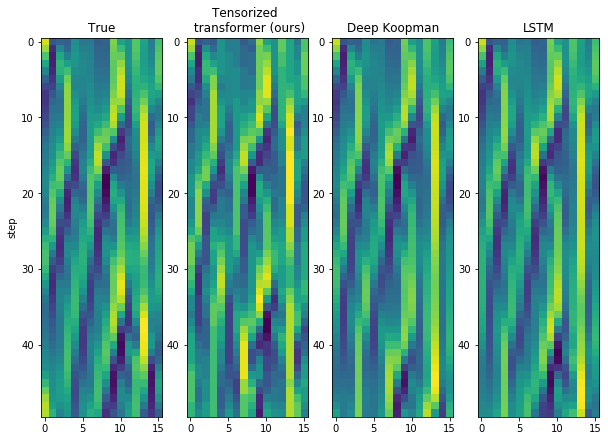}
    \caption{From left to right: real trajectory of the Lorenz-96 model with $d = 16$ and trajectories restored by Tensorized Transformer model with $N = 33$, Deep Koopman and LSTM models. Both LSTM and Transformer are conditioned on 100 states of the real trajectory.}
    \label{fig:lorenz}
\end{figure}

\subsection{Rossler Attractor}
\label{experiment:rossler}
is attractor for the \textbf{Rossler system}:
\begin{equation}\label{dynsys:rossler}
  \begin{split}
    \frac{dx}{dt} = -y - z, \
    \frac{dy}{dt} = x+ay, \
    \frac{dz}{dt} = b + z(x - c).
  \end{split}
\end{equation}
We use the system with $a=0.15, b=0.2, c=10.$ The training dataset consists of 1000 trajectories corresponding to the time $t = 1000$ with a time step $\tau = 0.1$. Initial states are sampled from distribution $x = x_0 + \epsilon$ with $x_0 = [5, 0, 0]$ and uniform $\epsilon \sim \mathrm{Uni}(-1, 1)$. For testing we use $20$ trajectories with the initial conditions samples from the same distribution. In this experiment, we use discretization size $N= 50$. The transformer model has 12 self-attention layers with $n= 9$ heads and size of representations $729$. 

Examples of the reconstructed trajectories are presented on Figure \ref{fig:rossler}; shown simulation corresponds to $t = 60$. All models capture general behaviour, still, performance of the transformer model is better, that corresponds to the Root Mean Squared Error (RMSE) that is demonstrated in Figure \ref{fig:rmse}. The graph represents dependence of the predictions $\tilde{x}(t)$ error from time:
\begin{equation}
    RMSE(\tilde{x}_i, x_i, t) = \sqrt{\frac{\sum_i \|x_i(t) - \tilde{x}_i(t)\|_2^2}{n}},
\end{equation}
averaged over $n = 100$ initial conditions. For both transformer and LSTM models, 100 steps were used as a conditioning. It is shown the GPT-2 model significantly outperforms regression models in the long-term perspective. After about 50 steps, the transformer model compensates inaccuracy of the single-state prediction and demonstrates much more stable behaviour then the Deep Koopman and LSTM models.

\subsection{Lorenz-96 System}
\label{experiment:lorenz}

We test scalability of the model on the example of \textbf{Lorenz-96} system with $d = 16$:

\begin{equation}
    \label{eq:lorenz96}
    \frac{dx_i}{dt} = (x_{i+1} - x_{i-2})x_{i-1} - x_i + F, \ i \in 1...d,
\end{equation}
where $x_{-1} = x_{d - 1}, \quad x_0 = x_d, \quad x_{d+1} = x_1$. We use forcing $F = 10$ and iteratively increase the grid size from $N = 2$ to $N = 33$ segments along every axis. Final grid corresponds to $V = 33^{16} \approx 2 \cdot 10^{24}$ size of the vocabulary. There are $1000$ and $20$ trajectories in train and test sets; each trajectory corresponds to the simulation for $20000$ steps with $\tau = 0.05$. Initial states are chosen as the vector $x_0 = F\cdot\mathbf{1}$, where perturbation $\epsilon \sim \mathcal{N}(0, 0.01)$ is added only to the $16$-th component of the vector. For Lorenz-96 systems we used transformer with 12 layers with $n = 8$ attention heads each and representations of size $1024$.

For this system, the exponential divergence of the real trajectories appears for the scale compared to the one of the discretization grid. As a result, the fact that we lose the micro-scale information is crucial for the restoration of the given trajectory. One can note that the discretization process is not bijective because the same sequence of multi-indexes corresponds to the infinite number of continuous trajectories. If the continuations of these trajectories are already uncorrelated, then the probabilistic model can not be expected to generate a precise approximation of a single solution. Instead, it samples continuation from all possible trajectories. In Figure \ref{fig:rmse} (right), the divergence of real trajectories is shown to be comparable with the behavior of the ones generated by the tensorized transformer. 

\begin{figure}
    \centering
    \includegraphics[width =0.9\textwidth]{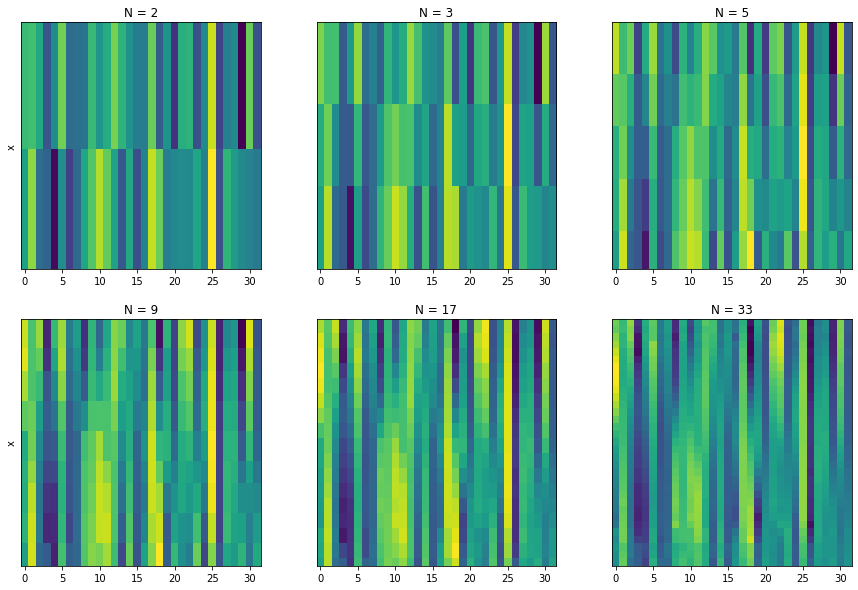}
    \caption{The first component of TT-embeddings (corresponds to $x_1$) after each iteration of fine-tuning; $N$ is the number of the discretization segments. Fine grid representations have the same pattern as the ones with a small number of segments. The representations get smoother with the growth of the discretization size $N$.}
    \label{fig:components}
\end{figure}

We also demonstrate the iterative training algorithm in action. In Figure \ref{fig:components}, there are the components of the representation $G_1$ corresponding to the $x_1$ coordinate of the Lorenz-96 system that is learned by the tensorized transformer after each iteration of the progressive training. After the first step, there are only two segments, so each component takes one of two possible values. On the next iteration, the axis is divided into three segments, then five and so on. Transitions to the finer grids are smooth, which proves that linear initialization is relevant in this case.

\section{Related Work}
Recently, deep neural networks have been replacing standard regression methods for dynamical systems modeling. Besides specific approaches like Deep Koopman model \cite{lusch2018deepkoopman, yeung2019deepkoopman} and Physics-Informed neural networks \cite{raissi2019physics, tipireddy2019physicsds, doan2019physicsreservoir, pangphysics}, various universal architectures including ResNet \cite{chashchin2019resnet}, ODEnet \cite{chen2018neural, rubanova2019latent, portwood2019turbulence, de2019gru} and recurrent models \cite{bailer1998recurrent, trischler2016synthesis, pan2018long, gers2002lstmlaser, vlachas2018lstm, yu2017ttrnn, vlachas2020backpropagation} have been applied to the dynamics modeling of chaotic systems. Still, none of the mentioned methods can preserve the invariant set of the dynamical system without additional modifications. One of the possible solutions was presented in \cite{trischler2016synthesis}; it was proposed to predict $x_{k+1}$ that belongs to the unitary cube and then re-scale it with a linear transformation. This approach bounds the predictions but does not hold the shape of the invariant set. An alternative solution is the usage of an additional model bounding the predictions if they have low probability according to the training set \cite{vlachas2018lstm}.

The radical alternative was proposed in \cite{shalova2020deep}; the authors suggested a probabilistic approach based on the transformer model. The equidistant grid discretization was used to map continuous states to the discrete representations. Then the model was trained to reproduce the conditional probabilities of the next state conditioned by the states' history. It was shown that by sacrificing a micro-scale data, it is possible to obtain a good approximation of the long-term system's behavior. The discretization of the phase space allowed modeling of the dynamics strictly holding the invariant set. One can note that a similar discrete approach to the conditional probabilities modeling is used in distributional RL \cite{bellemare2017distributional}. At the same time, in \cite{shalova2020deep} the geometrical properties of the phase space were not incorporated into the discrete-space model. As a result, proposed discretization led to the $O(N^d)$ size of the model.

\section{Discussion}
\textbf{Limitations.}
Although the tensorization allows the handling of larger systems, such an approach has several drawbacks. For high dimensions $d > 50$, even the iterative training approach does not lead to the efficient training of the full model, and many dynamical systems (such as discretizations of partial differential equations) have much higher dimensionality. Another restriction is a very simple factorized form of the probability distribution. Finally, it is not clear at the moment whether the developed method indeed learns the physics of the problem, and will it be able to predict something really "new'', and how dense should be the trajectories on the invariant set. 
%The factorized approach significantly broadens the applicability of the standard transformer model proposed in \cite{shalova2020deep}. 

\textbf{Future Work.} The factorization approach proposed in this paper does not rely on the architecture of the language model. In this work, we used a powerful transformer decoder model as a backbone; however, any of the NLP models are applicable. Despite the transformer being an undoubtable SOTA in discrete sequence modeling, the discrete probabilistic approach alone may give a significant improvement compared to the continuous space modeling. Another possible line of work is the incorporation of the developed pipeline into more complex models that include dimensionality reduction: the dynamical system is first mapped to a low-dimensional latent feature space, the discretized and modeled there. It would make the proposed approach more applicable for very large $d$.

\section{Conclusion}
%Regressive approaches do not guarantee reliable predictions of the chaotic dynamical systems due to the sensitivity to the initial conditions of the underlying equations.
In this paper, we introduced a probabilistic transformer-based model that learns the dynamics in the discrete space. To handle the dimensionality curse, we proposed using a tensor coding layer and factorizing the conditional distribution into the components corresponding to the system's dimensions. Together these approaches give $O(dN)$ estimation of the number of parameters. The model with fine discretization is shown to be superior to the LSTM and Deep Koopman models. Moreover, even in the case of the rough discretization scale, the model demonstrates interpretable behavior. 
%In addition, we demonstrated that geometrical relations make possible training of the transformer model with outstanding vocabulary size. 
Due to the complexity of the tensor coding layer, it requires a specific training method. For this purpose, we introduced a progressive training algorithm, that allows fine-grid modeling of dynamical systems. %such that model learns representations starting from the rough grid and iteratively increases the number of discretization segments $N$ on each iteration.

%\section{Broader Impact}
%This work does not present any foreseeable societal consequence.
%\bibliographystyle{unsrt}
%\bibliography{biblio}

\end{document}